%% file: main.tex
\documentclass[10pt,twocolumn,letterpaper]{article}

\usepackage{iccv}
\usepackage{times}
\usepackage{epsfig}
\usepackage{graphicx}
\usepackage{amsmath}
\usepackage{amssymb}

\usepackage{enumitem}
\DeclareMathOperator*{\argmin}{arg\,min}
\usepackage{algorithm}
\usepackage{algorithmic}
\usepackage{prettyref}
\usepackage{wrapfig}

\usepackage[pagebackref=true,breaklinks=true,letterpaper=true,colorlinks,bookmarks=false]{hyperref}

\iccvfinalcopy 

\def\iccvPaperID{2569} 

\ificcvfinal\fi
\begin{document}

%
\newcommand{\fch}[1]{{\textbf{\color{cyan}[FC: #1]}}}
\newcommand{\bki}[1]{{\textbf{\color{darkgreen}[BK: #1]}}}
\newcommand{\rme}[1]{{\textbf{\color{purple}[RM: #1]}}}
\newcommand{\jbr}[1]{{\textbf{\color{blue}[JB: #1]}}}

%
\title{PaintBot: A Reinforcement Learning Approach for Natural Media Painting}


\author{Biao Jia \\
Univsersity of Maryland \\
{\tt\small biao@cs.umd.edu}
\and
Chen Fang \\
ByteDance AI Lab\\
{\tt\small fangchen@bytedance.com}
\and
Jonathan Brandt \\
Adobe Research \\
{\tt\small jbrandt@adobe.com}
\and
Byungmoon Kim \\
Adobe Research \\
{\tt\small bmkim@adobe.com}
\and
Dinesh Manocha \\
University of Maryland \\
{\tt\small dm@cs.umd.edu}
}



%
\maketitle

\begin{abstract}
We propose a new automated digital painting framework, based on a painting agent trained through reinforcement learning.
To synthesize an image, the agent selects a sequence of continuous-valued actions representing primitive painting strokes, which are accumulated on a digital canvas.  Action selection is guided by a given reference image, which the agent attempts to replicate subject to the limitations of the action space and the agent's learned policy.  The painting agent policy is determined using a variant of proximal policy optimization reinforcement learning.  During training, our agent is presented with patches sampled from an ensemble of reference images.  To accelerate training convergence, we adopt a curriculum learning strategy, whereby reference patches are sampled according to how challenging they are using the current policy. We experiment with differing loss functions, including pixel-wise and perceptual loss, which have consequent differing effects on the learned policy.  
We demonstrate that our painting agent can learn an effective policy with a high dimensional continuous action space comprising pen pressure, width, tilt, and color, for a variety of painting styles. Through a coarse-to-fine refinement process our agent can paint arbitrarily complex images in the desired style.
\end{abstract}

\input{intro.tex}
\input{related.tex}
\input{overview.tex}

\input{method.tex}
\input{result.tex}
\input{conclusion.tex}

{
\bibliographystyle{ieee}
\bibliography{main}
}


\end{document}

%% file: intro.tex
\section{Introduction}

\label{sec:intro}
Throughout human history, painting has been an essential element of human culture, and has evolved to become a massively diverse and complex artistic domain, comprising thousands of different styles, including subtle watercolor scenes, intricate Chinese ink landscapes, and detailed oil portraits of the Dutch masters. 
Over the last few decades, there has been considerable effort to simulate some of these styles through non-photorealistic rendering techniques, including stroke-based rendering and painterly rendering techniques using manually engineered algorithms~\cite{hertzmann1998painterly, winkenbach1996rendering}.  These efforts have produced compelling results, but are hampered by their dependency on hand-engineering to produce dramatically new styles.

\begin{figure}[t]
\centering
\includegraphics[width=0.45\textwidth]{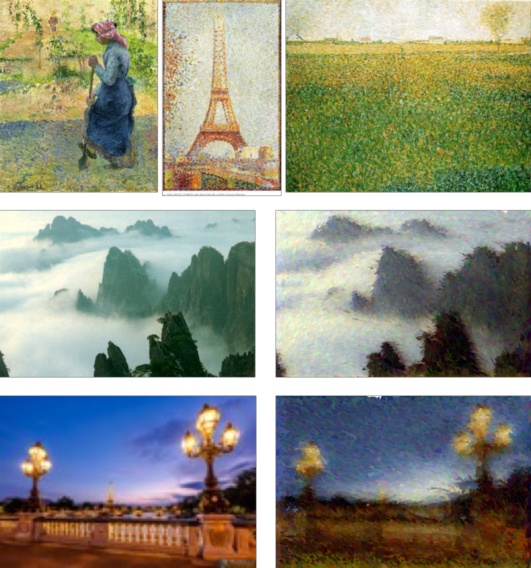}
\caption{\emph{Style Emulation using PaintBot:} 
We use three paintings (top row) of the Pointillism style as the training dataset for our reinforcement learning algorithm. Using as reference the images shown in the left column of the middle and bottom rows, PaintBot automatically produces the digitally painted images in the right column of the corresponding row.
}
\label{fig:pissaro}
\vspace*{-0.12in}
\end{figure}

Recent developments in machine learning have resulted in significant advancements in image recognition and image synthesis. Many of these learning methods have also been used for painting related tasks, including modeling the brush~\cite{xie2012}, generating brush stroke paintings in an artist's style ~\cite{xie2015stroke}, reconstructing drawings for specific style paintings ~\cite{tang2018animated}, and constructing stroke-based drawings~\cite{DBLP:journals/corr/HaE17}. Other approaches  are based on generative adversarial networks \cite{goodfellow2014generative} and variational autoencoders~\cite{kingma2013auto} and they have also  been applied to artistic painting styles \cite{zhu2017unpaired, zhou2018learning, DBLP:journals/corr/abs-1803-04469, karras2017progressive, sangkloy2017scribbler}.

In this paper, we focus on a more general and challenging problem of training a natural media painting agent from scratch using reinforcement learning methods. 
Our goal is to develop an agent that is able to perform a sequence of primitive drawing actions to produce a target output. We adopt a purely data-driven approach so that our method can be generalized to new painting styles by expanding the training set. 


 We present a novel automated digital painting framework that uses a painting agent trained through reinforcement learning for nature media painting.
In particular, our agent generates strokes step-by-step by issuing actions to a digital painting system, which we call the {\it Simplified Simulated Painting Environment} (SSPE).
Given a reference image, our painting agent aims to reproduce the identical or transformed version of that image in the SSPE. 
A particular challenge that arises in this framework, as compared to previous work, is the relatively large action space and the large number of steps that the agent must take to reach its goal.

We present a reinforcement learning-based framework to train the painting agent without human supervision. We adopt proximal policy optimization (PPO) \cite{schulman2017proximal} evaluate the visual quality of the resulting paintings.
Since our agent operates in a large continuous action space, we use techniques based on difficulty-based sampling and curriculum learning to make the training more efficient. 

The novel contributions of our work include:

\begin{itemize}[noitemsep,nolistsep]
    \item A novel deep reinforcement learning network that is designed for this particular task. Our approach learns without human supervision, works well in continuous high dimensional action space and does not degrade after thousands of strokes which can handle a large dense reference image. 
    \item To reduce the search space of the policy network, we design several techniques like curriculum learning and  difficulty-based sampling to improve the performance by $20\%$.
\end{itemize}

We have evaluated our results on a number of reference images with different styles as shown in Figure \ref{fig:pissaro}. Comparing with the visual generative methods \cite{hertzmann2001image, zhu2017unpaired}, Our results are promising and show that our painting agent can generate results of high resolution and can be applied to different painting media. The training phase takes about six hours and the runtime algorithm takes about $300$ seconds on each reference image on a desktop PC. 

%% file: related.tex
\section{Related Work}

\subsection{Stroke-based Painterly Rendering}
Our approach can be regarded as a type of stroke-based painterly rendering approach. These types of methods render an input image on canvas as a combination of strokes, and the designed algorithms determine the properties of such strokes, including position, density, orientation, length, width, color and so on. Hertzmann et al.~\cite{hertzmann1998painterly} propose an algorithm to render an input image into a painting with primitive strokes with controllable properties.  The algorithm is shown to be able to render various visual styles by combining different types of strokes/brushes. To simulate mosaic decorative tile effects, Hausner et al.~\cite{hausner2001simulating} design an algorithm based on  Centroidal Voronoi diagrams that can place square tiles and respects the gradient and color of a reference image. Generally speaking, it is necessary to hand engineer new algorithms for new styles such as stipple drawings~\cite{deussen2000floating}, pen-and-ink sketches~\cite{salisbury1994interactive} and oil paintings~\cite{zeng2009image} ~\cite{lindemeier2015hardware}. In contrast to these approaches, our method automatically learns to compose the selected style, therefore avoiding the expensive manual design process.

\subsection{Learning-based Drawing}
There have been a few attempts to tackle related problems in this domain. Xie et al.~\cite{xie2012,xie2015stroke,xie2013personal} propose a series of works to simulate strokes using reinforcement learning and inverse reinforcement learning. The proposed approaches learns a policy either from reward functions or expert demonstrations. Unlike our goal, Xie et al.~\cite{xie2012,xie2015stroke,xie2013personal} mainly focus on designing reward functions for generating oriental painting strokes and their method requires expert demonstrations as supervision. Recently, Ha et al.~\cite{DBLP:journals/corr/HaE17} collected a large-scale dataset of millions of simple sketches of common objects with the corresponding recording of painting actions. Based on this dataset, a recurrent neural network model is trained in a supervised manner to encode and re-synthesize action sequences, and the trained model is shown to be capable of generating new sketches. Following \cite{DBLP:journals/corr/HaE17}, Zhou et al.~\cite{zhou2018learning} exploits reinforcement learning and imitation learning to reduce the amount of supervision needed to train such a sketch generation model. Distinct from \cite{DBLP:journals/corr/HaE17,zhou2018learning}, our painting agent operates in a complex SSPE with a much larger action space, including pen pressure, tilt and color, and our approach learns its policy network completely without human supervision.

\subsection{Visual Generative Methods}
Visual generative methods typically directly synthesize visual output in pixel spaces, which is fundamentally distinct from our approach. 
Image analogies \cite{hertzmann2001image} solve the problem by introducing a non-parametric texture model.  More recent approaches based on CNNs using large dataset of input-output training image pairs
to learn the mapping function \cite{gatys2015neural}. 
Inspired by the variaonational auto-enhnson et al.\cite{johnson2016perceptual}
Inspired by the idea of variational autoencoders \cite{kingma2013auto}, 
Johnson et al. \cite{johnson2016perceptual} introduce the concept of perceptual loss to implement style transferring between paired dataset. 
Inspired by the idea of generative adversarial networks \cite{goodfellow2014generative},  
Zhu et al. ~\cite{zhu2017unpaired} learn the mapping without paired training examples using Cycle-Consistent Adversarial Networks. 
Such methods have been successfull at generating natural images~\cite{karras2017progressive,sangkloy2017scribbler}, artistic images~\cite{li2017universal} and videos~\cite{vondrick2016generating,li2018flow}.
In terms of the final rendering, the current visual generative methods can produce result in various painting styles using limited traing dataset. Comparing with our method, these generative methods may fail to get results of high resolution. For the purpose of interactive artistic creation, stroke-based approach can generate trajectories and intermediate painting state. Another advantage of stroke-based method is that the final results are trajectories of paint brush, which can be deployed in different synthetic natural media painting environment and real painting environment using robot arms.

\subsection{Reinforcement Learning}
Reinforcement learning (RL) has achieved promising results recently in many problems, such as playing Atari games~\cite{mnih2013playing}, the game of Go~\cite{silver2017mastering} and robot control~\cite{levine2016end}. A major focus of this effort has been to achieve improved time and data efficiency of the learning algorithms. Deep Q-Learning has been shown to be effective for tasks with discrete action spaces~\cite{mnih2013playing}, and proximal policy optimization (PPO)~\cite{schulman2017proximal} is currently regarded as one of the most effective for continuous action space tasks. For the painting task, we require a continuous action space to capture brush movement and color variation.  We therefore adopt PPO as the basis for our learning method.

Intrinsic to any RL problem is a well-defined reward function that guides policy learning.  For instance, the reward in a game setting might be number of wins versus losses or points gathered. However, in painting the reward is not clearly nor uniquely defined, as evidenced by the myriad diverse yet individually appealing visual styles that human artists have developed. In this work, we have explored the choice of a reward function, in addition to the learning algorithm.  In particular, we have explored both L2 and perceptual distance to a given reference image.  The fact that the reference image can be selected arbitrarily for any given roll-out has presented challenges to the vanilla PPO framework.  We address these challenges through curriculum learning technique, as will be described below.






\nocite{zheng2018strokenet,tang2018animated,xie2015stroke,xie2012,DBLP:journals/corr/HaE17}

%% file: overview.tex
\begin{figure}[t]
\centering
\includegraphics[width=0.45\textwidth]{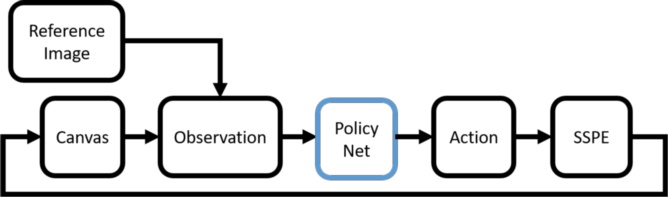}
\caption{{\em Overview of PaintBot Operation:} For each time step of the painting simulation, the current state of the canvas and the reference image form the observation for the policy network. Based on the observation, the policy network selects an action for the painting engine (SSPE) to execute and update the canvas accordingly.}
\label{fig:apply}
\vspace*{-0.15in}
\end{figure}

\section{Notation and Overview}
\label{sec:overview}
In this section, we introduce the notation used in the paper and give an overview of our approach.

\begin{table}[t]
\begin{tabular}{ll}
\hline  
Symbol & Meaning   \\
\hline
$s$ & current painting state	\\
$s^*$ & target painting state \\
$o(s)$ & observation the painting state \\
$r$ & reward \\
$\pi$ & painting policy, predict $a$ by $o$\\
$V_\pi$ & value function of the painting policy, \\
& predict $r$ by $o$ \\
$L$ & loss function, measuring distance between \\
    & observations\\
$I_{ref}$ & reference image \\
$\{I^{(t)}_{ref}\}$ & set of reference images \\
$I_i$ & canvas of $i_{th}$ time step \\
$p_i$ & position of the paint brush of $i_{th}$ time step \\
$a_i$ & action of $i_{th}$ time step \\
$\alpha$ & angle of the stroke	\\
$l$ & stroke length	\\
$c$ & stroke color	\\
$w$ & stroke width \\
\hline
\end{tabular}
\caption{\label{Fig:param} Notation Summary}
\vspace*{-0.20in}
\end{table}

\subsection{Training the PaintBot}
We demonstrate the overall approach using the flowchart in Fig.\ref{fig:apply} and illustrate the training and test procedure in a step-wise order. 
Our approach adopts proximal policy optimization \cite{schulman2017proximal} to train the model by sampling actions at each step. The policy function is implemented using a deep neural network. 
 Training via very long episodes would result in a very large number of samples and hence extremely slow training. Therefore, we use a series of methods to reduce the amount of sampling process and accelerate the training. First, we set a limitation on the maximum steps for each trial, even if the agent fails to achieves the goal configuration. Next, we increase the limit gradually. We use curriculum learning (Section ~\ref{sec:curriculum}) to encourage the agent to find the reward greedily in limited time steps to reduce the  possible exploration space. 
Second, we incorporate difficulty-based data sampling (Section~\ref{sec:difficulty}) to overcome the bias between different samples. 
For the common reinforcement learning tasks, the goal configuration is usually fixed. In our case, the reference image  $I_{ref}$ has to sample from a set of images $\{I^{(t)}_{ref}\}$ to prevent over-fitting. Third, we replace the $L_2$ loss with $L_{\mbox{\scriptsize percept}}$ and $L_{\frac{1}{2}}$, to give more rewards for matching the color and shape of a reference image exactly than for finding an average color. 
$L_{\mbox{\scriptsize percept}}$ and $L_{\frac{1}{2}}$ are defined in Sec.\ref{sec:reward}.

\subsection{Roll-out the PaintBot }
As Fig.\ref{fig:apply} shows, we apply the trained policy to render the final result. First, we feed the observation to the policy networks and compute the output action. After rendering the action, we fetch the updated observation. The action is defined as a continuous vector of stroke configurations composed of  angle, length, width, and color. The visual part of the observation consists of the reference image and current canvas. We take a patch centered at the current pen location from the reference and the current canvas, rather than from the entire image. We call this egocentric observation. This observation helps the network to attend more to the neighborhood of the current pen location.

We apply the policy at different image scales of the reference image. As shown in Fig.\ref{fig:scale}, we let the agent start with a low resolution reference image in order to quickly paint the high-level visual structure, and gradually move it to a high resolution reference image to let the agent draw details. Note that, later scales paint directly on the canvas from previous scales. 

\begin{figure}[t]
\centering
\includegraphics[width=0.45\textwidth]{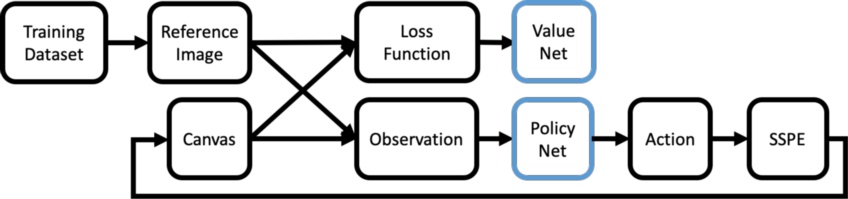}
\caption{{\em Training framework}:  (1) We sample a reference image as the goal configuration from the training dataset;
(2) We feed the loss function with the goal configuration and canvas to compute the reward of the current state;
(3) We concatenate the egocentric patches of reference image and canvas as the observation;
(4) We feed the reward to the value network $V_\pi$ and the observation to the policy network $\pi$;
(5) The policy network predicts the action by the observation;
(6) SSPE performs the action and renders the updated canvas.}
\label{fig:overview}
\vspace*{-0.15in}
\end{figure}

%% file: method.tex
\section{Painting Agent}
In this section, we present technical details of our painting agent that is based on reinforcement learning. First, we introduce the basic setup of the reinforcement learning, including the definition of action space, observation, reward, and policy network. Afterwards, we explain the details of our training algorithm and run-time algorithm, and also some methods to improve the learning efficiency like curriculum learning and imitation learning.

\subsection{Policy Representation}
The policy of the painting agent includes the definition of action, observation, reward, and the structure of the policy network. The action space denotes the degrees of freedoms of the painting agent, which is the input of the policy network. The observation denotes the state of the painting process, which is the input of the policy network. The reward function measures the advantage of the painting action towards the goal configuration, which is given by the environment. The structure of the policy network defines the technical implementation of the machine learning approach.

\begin{figure}[t]
\centering
\includegraphics[width=0.2\textwidth]{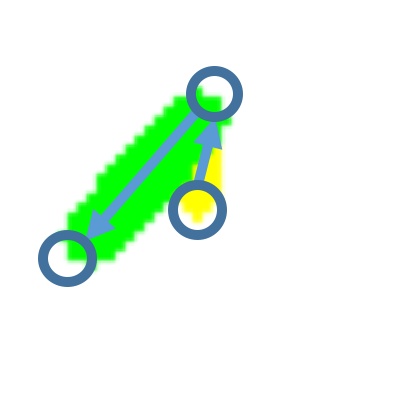}
\put(-40,90){$p_{i-1}$}
\put(-80, 60){$a_i$}
\put(-105,25){$p_i$}
\put(-35, 60){$a_{i-1}$}
\put(-40,40){$p_{i-2}$}
\put(-105, 10){$a_i=\{\alpha_i, l_i, c_i\}$}
\put(-105, 1){$a_{i-1}=\{\alpha_{i-1}, l_{i-1}, c_{i-1}\}$}
\caption{{\em Action Representation:} Given $p_{i-1}$, the position of the $i-1$th step, and $a_i$, , the action of $i$th step,  we have $p_i = p_{i-1} + [l_{i-1}sin(\alpha_{i}), l_{i-1}cos(\alpha_{i})]$}. 
\label{fig:action}
\vspace*{-0.15in}
\end{figure}
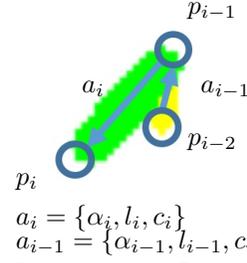


\subsubsection{Action Space}
\label{sec:action}
To highlight the painting behavior, we denote the action using properties of a stroke, including position, size, and color. Specifically,  we define the action as a 6-dimensional vector, $a_i=[\alpha, l, w, c_r, c_g, c_b] \in \mathbb{R}^6$.  Each value is normalized to $[0, 1]$. The action is in a continuous space, which makes it possible to train the agent using policy gradient based reinforcement learning algorithms. 
Specifically, when $w=0$, the brush does not paint on the canvas but moves above the canvas
to the updated position. The action representation and the computation between the actions and position is shown as Fig.\ref{fig:action}.


\subsubsection{Observation}
\label{sec:obs}
For most reinforcement learning setups, the goal state of each training episode is always the same although the initial state of the each training episode may vary. Thus the setup of these problems does not code the goal state as part of the observation, and the model will implicitly learn the fixed goal state.   

A painting state $s_i$ is usually considered as two parts, the canvas and the position of the paint brush,  defined as $s_i = \{I_i, p_i\}$. 
To generalize the model for different reference images, the training approach should consider a goal state, which is the reference image $I_{ref}$, as a part of the observation as $s_i = \{I_i, I_{ref}, p_i\}$. For all our experiments, we code both the reference image and the canvas as the observation to describe the current state and the goal state of the agent.

Another challenge is to incorporate the positional information of the paint brush
into the observation. We use egocentric observation, which means we always keep the paint brush in the center of the canvas, and move the canvas and reference image instead. 
Thus, egocentric observation does not use the position, and is  position-agnostic. This approach greatly reduces action space greatly and does not require a replay buffer. Thus egocentric observation renders the challenging problem of training an agent in the continuous action space and large state space is a tractable problem.
The observation of state $s_i$ is defined as Eq.\ref{eq:obs}, where $(p_h,p_w)$ is the 2D position of the paint brush, $(h_o, w_o)$ is the size of the egocentric window.

\begin{equation}
\label{eq:obs}
\begin{split}
    o(s_i) = &\left\{ I_i\left[p_h-\frac{h_o}{2}:p_h+\frac{h_o}{2}, p_w-\frac{w_o}{2}:p_w+\frac{w_o}{2}\right]\right.,   \\      &\left.I_{ref}\left[p_h-\frac{h_o}{2}:p_h+\frac{h_o}{2}, p_w-\frac{w_o}{2}:p_w+\frac{w_o}{2}\right] \right\}.
\end{split}
\end{equation}

By defining the observation, we can incorporate the position of the paint brush and training data can be generalized. When the training process proceeds, the training data $\{o(s_i), a_i, r_i\}$ can been seen as many sampled patches from the original reference image.

\begin{figure}[t]
\centering
\includegraphics[width=0.4\textwidth]{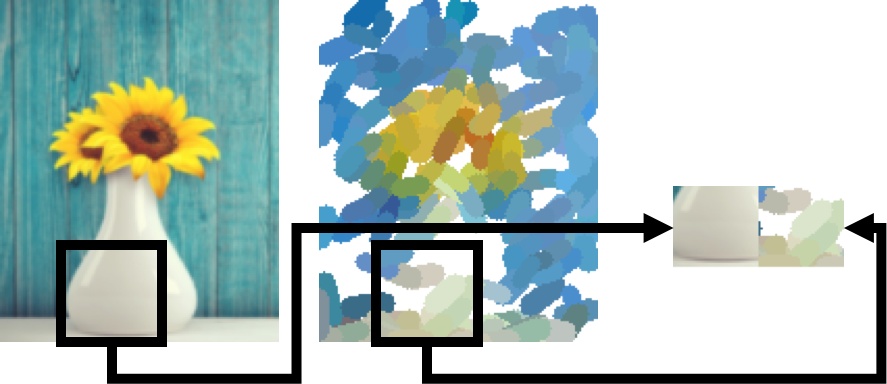}
\caption{The observation for the Painting Agent is formed by concatenating two egocentric patches of the reference image and current canvas, respectively.
}
\label{fig:observation}
\end{figure}
\vspace{-0.1in}

\subsubsection{Reward}
\label{sec:reward}
In our setup, the reward of each action is defined by the difference between the canvas and the reference image. 
A loss function is applied to compute the reward of an action during each iteration of the reinforcement learning.
To better reproduce the reference image, the reward should stimulate the agent to paint strokes to reduce the distance between the current canvas and the reference image as much as possible. 

We use different loss functions to get the best visual effects of the final renderings. 
The original $L_2$ loss can be formulated as:
\begin{equation}
    L_2(I, I^{ref}) = \frac{\sum^h_{i=1}\sum^w_{j=1}\sum^c_{k=1}||I_{ijk}-I^{ref}_{ijk}||_2^2}{hwc}
\end{equation}
where the image $I$ and the reference image $I_{ref}$ is a matrix whose shape is $h \times w \times c$. In this case, $w,h$ are width and height of the image, and $c$ is the number of color channels.

To encourage the painting agent to match the color and shape of the reference image exactly rather than to finding an average color, we modify the $L_2$ loss into $L_{\frac{1}{2}}$.

\begin{equation}
     L_{\frac{1}{2}}(I, I^{ref}) = \frac{\sum^h_{i=1}\sum^w_{j=1}\sum^c_{k=1}|I_{ijk}-I^{ref}_{ijk}|^{\frac{1}{2}}}{hwc}
\end{equation}

\begin{figure}[t]
\centering
\includegraphics[width=0.4\textwidth]{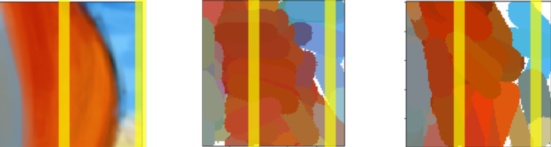}
\put(-185,-10){(a)}
\put(-105,-10){(b)}
\put(-30,-10){(c)}
\caption{\emph{Loss Function Comparison:} We compare the visual effects of differing loss functions: (a) reference image, (b) painting result using $L_2$ loss function, (c) painting result using $L_{\frac{1}{2}}$ loss function. 
(The most noticeable differences are highlighted with the yellow box.) 
In general, agents trained with $L_{\frac{1}{2}}$ tends respect the shape boundaries better, while ones trained with $L_2$ loss tend to match the average color better.)
}
\vspace*{-0.2in}
\end{figure}

We also use a perceptual loss-based~\cite{johnson2016perceptual} reward to encourage the agent to have similar feature representations similar to those computed by the loss network $\phi$.  
In this case, the loss network $\phi$ is a convolutional neural network for classification purposes, like \cite{vgg16} and \cite{resnet}.
It is implemented by comparing the Euclidean distance of the rendered image and the reference image between the feature representations:
\begin{equation}
\label{eq:percept}
L_{\mbox{\scriptsize percept}}(I, I^{ref})= \sum^N_{n=1} \frac{||\phi_n(I)-\phi_n(I^{ref})||_2^2}{h_n w_n c_n},
\end{equation}
where the shape of the feature map of $\phi_n$ is $h_n \times w_n \times c_n$. 
After we define the loss between $I$ and $I^{ref}$, we normalize $r_i$ using Eq.\ref{eq:reward}, such that $r_i \in (-\infty, 1]$.
\begin{equation}
\label{eq:reward}
r_i = \frac{L(I_{i-1}, I_{ref}) - L(I_{i}, I_{ref})}{ L(I_{0}, I_{ref})},   
\end{equation}
where L is a loss function defined above as, $L_2$,  $L_{\frac{1}{2}}$ or $L_{\mbox{\scriptsize percept}}$.


\subsubsection{Policy Network}
To define the structure of the policy network, we consider the input as a concatenated patch of the reference image and canvas $41 \times 82 \times 3$ given the sample size of $41 \times 41 \times 3$. 
The first hidden layer convolves 64 $8 \times 8$ filters with stride 4, the second convolves 64 $4 \times 4$ filters with stride 2 and the third layer convolves 64 $3 \times 3$ filters with stride 1. After that, it connects to a fully-connected layer with 512 neurons. All layers use ReLU activation function \cite{krizhevsky2012imagenet}.

\begin{figure}[t]
\centering
\includegraphics[width=0.4\textwidth]{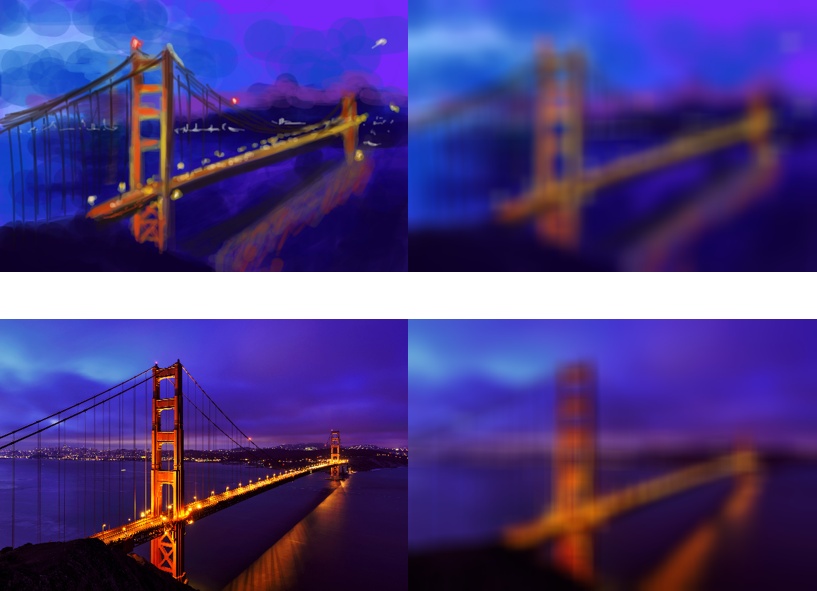}
\put(-55,72){(b)}
\put(-55,-10){(d)}
\put(-155,72){(a)}
\put(-155,-10){(c)}
\caption{\emph{Blurred Observation for Style Transfer:} We perform Gaussian blur \cite{chan1998total} to both target image (a) and reference image (c) to get corresponding blurred images (b) and (d). 
The two images are more similar after low-pass filtering, which reduces the loss defined in Eq. \ref{eq:reward} and consequently eases the style transfer task.}
\label{fig:blur}
\vspace*{-0.12in}
\end{figure}
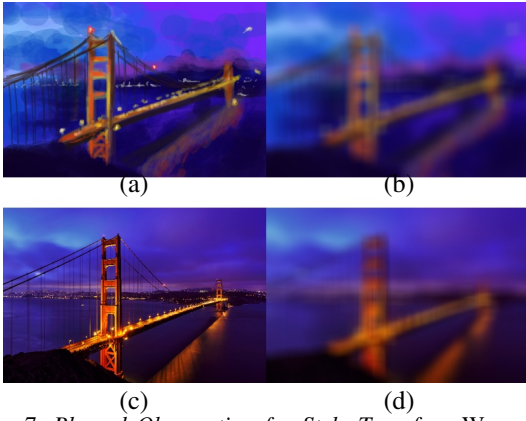

\subsection{Reinforcement Learning}
\begin{algorithm}[t]
  \caption{Difficulty-based Sampling 
}
  \label{alg:sampling}
  \begin{algorithmic}[1]
    \REQUIRE 
     Reference images $\{I^{(t)}_{ref}\}$ with sampling amount $n$, total iterations $N$
    \ENSURE Painting Policy $\pi$ and its value function $V_\pi$
    \STATE $\{r^{(t)}\}$ // Mean reward tested using the sample 
    \FOR {$i = 1, \cdots, t_{max}$}
    	\STATE $r_i$ = 0 //Initialization
    \ENDFOR
    
    \FOR {$iter = 1, \cdots, N$}
        \FOR {$i = 1, \cdots, n$}
        	\STATE $r_i$ = $V_\pi(o(I^{(i)}_{ref}))$  // initialize the reward with the policy 
        	\IF {$r_i<r_{min}$}
        	    \STATE $r_{min}$ = $r_i$ 
        	    \STATE $min_i$ = $i$
        	\ENDIF 
        \ENDFOR

        \STATE $\pi=\textsc{Update}(\pi, o(I^{(min_i)}_{ref}))$
    \ENDFOR
   \RETURN $\pi$ 
  \end{algorithmic}
\end{algorithm}

\begin{algorithm}[t]
  \caption{Run-time Algorithm}
  \label{alg:test}
  \begin{algorithmic}[1]
    \REQUIRE Reference image $I$ which size is $(h_I, w_I)$, \\
    the learned painting policy $\pi$ with observation size $(h_o, w_o)$
    \ENSURE Final rendering $I^*$
    
    \WHILE {$||I-I^*||>Thresh_{sim}$ }
        \STATE $p=(random(h_I), random(w_I))$ //sample a 2-dimensional point within image to start the stroke $I$
        \STATE $o = I[p_0-\frac{h_o}{2}:p_0+\frac{h_o}{2}, p_1-\frac{w_o}{2}:p_1+\frac{w_o}{2}]$ //Get observation
        \STATE $r = 1$ //Initialize the predicted reward
        \WHILE { $r>0$}
            \STATE $\alpha, l, c, w = \pi(o)$ //Predict the painting action
            \STATE $r = V_\pi(o)$            //Predict the expected reward
            \STATE $I = R(I, p, \alpha, l, c, w)$ // Render the action
            \STATE $p = (p_0+ l \times cos(\alpha), p_1+l \times sin(\alpha))$ //Update the stroke position
            \STATE $o = I[p_0-\frac{h_o}{2}:p_0+\frac{h_o}{2}, p_1-\frac{w_o}{2}:p_1+\frac{w_o}{2}]$ //Update the observation
        \ENDWHILE
    \ENDWHILE
    \RETURN $I$
  \end{algorithmic}
\end{algorithm}

\subsubsection{Curriculum Learning}
\label{sec:curriculum}
Due to the continuous action space $a \in \mathbb{R}^6$, the sampling space can be extremely large as the number of time steps increasing. Further the signal can be overwhelmed by the noise while applying policy gradient based reinforcement learning algorithms.
To train the model efficiently,  we train the agent in a curriculum learning style, which means the sampled trajectories of the agents increase during training episodes. 
As a result, the agent can stage learned policy and generate relatively long strokes compared to the model trained without the techniques. The agent tends to find the reward greedily in the limited time steps.

Another main challenge is the bias between different samples; for the common RL tasks, the goal is usually fixed. In our case, however, the reference image must change to prevent an over-fitting problem.
To overcome this, we incorporate difficulty-based data sampling. 
In reinforcement learning, the optimal policy $\pi^*$ maximizes the expected long term reward $J(\pi)$, which is accumulated by discounted rewards $r_i$ in a horizon $t_{max}$ of steps with a factor $\gamma \in \mathbb{R}$,
\begin{equation}
    J(\pi) = \sum^{t_{max}}_{t=1}{r_t\gamma^t},
\end{equation}
where $t_{max} \in \mathbb{Z}$ is usually fixed as the maximal number of steps for each trial.  

For a painting policy, there are a lot of goal configurations which are distributed sparsely in a high dimensional space, which can make the converging process fail because the agent can hardly compute the gradient of the policy. We modify the horizon parameter $t_{max}$ by introducing a reward threshold $r_{\mbox{\scriptsize thresh}}$, and make it increase gradually during the training process as: 
\begin{equation}
    \hat{t}_{\mbox{\scriptsize max}} = \argmin_i(r_i>r_{\mbox{\scriptsize thresh}}).
\end{equation}
Given the redefined horizon parameter, the policy gradient algorithm can converge efficiently with a set of complex goal configurations. The policy is encouraged to find rewards greedily in limited time steps to reduce the possible exploration space. 

\subsubsection{Difficulty-based Sampling}
\label{sec:difficulty}
As illustrated in Alg.\ref{alg:sampling}, we incorporate difficulty-based sampling method to select a goal configuration for each painting trial from a set of reference images. This sampling method can overcome the bias between different samples. 
For common reinforcement learning tasks, the goal is usually fixed. In our case, the reference image must change to prevent over-fitting. For each run of the agents, the environment should be initialized by  $p_0$ and $I^{(t)}_{ref}$, and $I^{(t)}_{ref}$ is selected from the training dataset $\{I^{(t)}_{ref}\}$ with size $n$.

For different $I^{(t)}_{ref}$, the maximum reward collected each run can vary throughout the training process when $t \in \mathbb{Z}$ is randomly sampled in $[0,n]$. Thus, the learning progress is not balanced among the dataset, which can cause the policy over-fitting for specific inputs. 
The difficulty-based sampling method is designed to mitigate the learning progress among the set of goal configurations. It encourages the approach to sample more from the images with the worst performances.

\subsubsection{Generalization}
Because the policy trained by reinforcement learning usually uses a fixed goal configuration, the most challenging problem is the generalization.
For our approach, it is crucial to add variation by defining a more generalizable loss, observation, and data preparation method. 

To make the learned policy to apply to reference images at different scales, we take the sliding window as the observation with an egocentric view. The variation of the observation greatly improves compared with the observation from a fixed view of the reference image. Moreover, we reduce the redundant information of the goal configuration by applying Gaussian blurring~\cite{chan1998total} to the reference image as the observation, while the loss function stays the same as defined in Section \ref{sec:reward} (see Fig.\ref{fig:blur}).
As a result, the goal is switched to reproduce the original image with the blurred input
\begin{equation}
\pi^*(blur(I)) = \argmin_{\pi}L(I, I_{ref}).
\end{equation}

To apply the policy to a reference image, we follow the algorithm described in Fig. \ref{alg:test}. We randomly select a point at which the agent will start a stroke, and then extract the local patches of the reference image and canvas as the observation.
Then, we iteratively apply the actions predicted  by the policy network to the canvas, until the value network predicts a small reward, which suggests the low expectation of the next action.
We also using the roll-out result as the initial canvas for training process, which aims to make the painting policy more robust to recover from the deviated state.
To model the different natural media, we also fine-tune our model using various painting media by modifying the environment parameters.

\begin{figure}[t]
\centering
\includegraphics[width=0.45\textwidth]{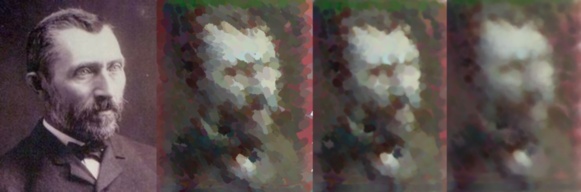}
\put(-200,-10){(a)}
\put(-140,-10){(b)}
\put(-80,-10){(c)}
\put(-30,-10){(d)}
\caption{\emph{Generalization using different painting media:} The figure shows the different rendering result using different painting media. For this case, we vary the color blending parameter of the stroke and the canvas. (a) reference image (b, c, d)  environment with color blending parameter $1.0, 0.6, 0.3$
.}
\label{fig:blur}
\vspace*{-0.15in}
\end{figure}
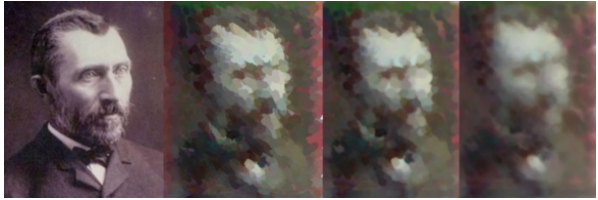

%% file: result.tex
\section{Implementation and Results}

In this section, we describe our implementation and highlight the performance of our painting agent on different benchmarks.

\subsection{Setup}
For our setup, we mainly use an environment in which painting agent can explore a high dimensional action space and observation space. To initialize the environment, the reference image and the canvas are given as  inputs. The reference image denotes the target configuration and the canvas denotes the initial configuration. 

For each running episode of the training process, the initial position of the agent is set at the center of the canvas. After that, our training algorithm runs in an iterative manner, as shown in Fig.\ref{fig:apply}, for $\hat{t}_{max}$ steps. 
This formulation of the environment works for cases with a single configuration. To train the model with different configurations, we use the difficulty-based sampling highlighted in Alg.\ref{alg:sampling} and initialize the environment with the configuration with the minimal mean reward.

\subsection{Data Preparation}
For the painting agent trained using the supervised learning method, we  need to use observation and action pairs as the training dataset. In practice, this requires a great amount of effort to collect such data from  human participants.
In our reinforcement learning framework, the actions corresponding to each observation are explored by the policy neural network. The painting model is learned implicitly by the reference images.
Thus, different painting styles are potentially learned by different sets of images.

\begin{figure}[t]
\centering
\includegraphics[width=0.38\textwidth]{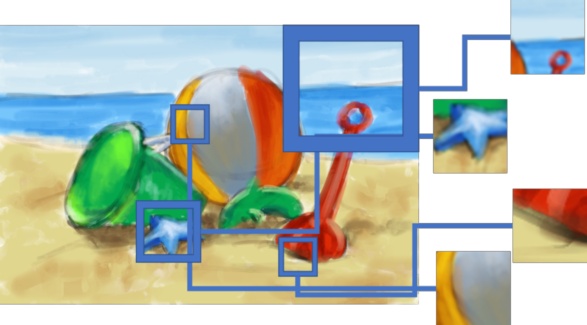}
\caption{\emph{Data Preparation:} To prepare the training data, we draw random patches from the reference image at varying scales, apply random rotation and flipping operations, followed by resampling the patches to a fixed size. }
\label{fig:data}
\vspace*{-0.12in}
\end{figure}

To apply our model at different scales of the reference image, we build a pyramid on the reference image and sample patches from it that are then normalized to a fixed shape to assemble the data set, shown as Fig. \ref{fig:data}. 
It is important to also ensure that the input images contain a sufficient variation in texture, color, and shape. Otherwise, it can make the painting policy subject to over-fitting. We cluster the final training samples using the perceptual loss metric defined in Eq.\ref{eq:percept}, and then choose a random sample in each cluster of the dataset.

In some cases, we also make use of the biased model to generate a specific style of painting. To do this, we select images in the same painting style or from one specific artist. After that, the biased model can generate the result in specific style, as shown as Fig.\ref{fig:style}.

\subsection{Performance}
\begin{figure}[ht]
\centering
\includegraphics[width=0.4\textwidth]{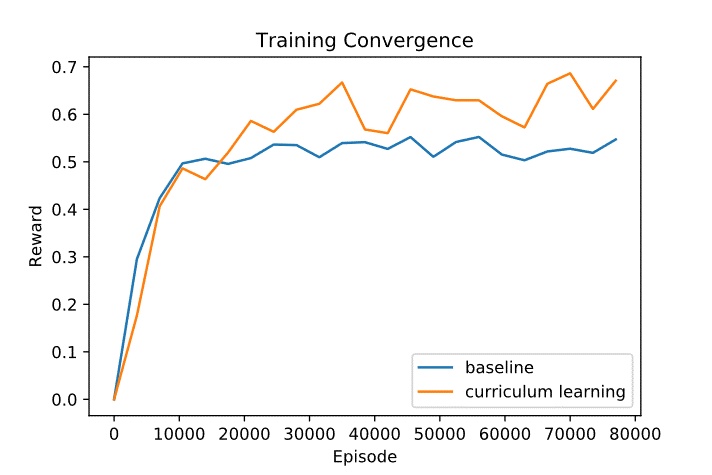}
\caption{\emph{Curriculum Learning} This figure compares the learning curve between approach using curriculum learning and baseline. The y-axis denotes the average rewards of the trained model in a validation dataset, and the x-axis denotes the training episodes. Both approaches converge after a certain number of steps, but the approach with curriculum learning behaves better with a higher reward value. 
The total training steps used in both approaches is about $10^6$.}
\label{fig:learning_curve}
\vspace{-0.1in}
\end{figure}

Figure \ref{fig:learning_curve} illustrate the learning curve of our algorithm comparing the baseline model and the model applied curriculum learning. Both models converge within $78,000$ episodes. The y-axis denotes the average rewards of the trained model in a validation dataset, and the x-axis denotes the training episodes. As the training process proceeds, the average rewards grows.

The overall training includes $10^6$ steps, which can be finished in 6 hours with an Intel i7 CPU and a GTX 1080 GPU. The training process can be shortened if  we increase the number of GPUs. 
After the training period, the application model is quite fast. 
The complexity of the run-time algorithm  is $O(h_I w_I h_o w_o)$, which is a linear function of the  
total number of pixels of the reference image and the observation patch. 
It takes about $300 secs$ to compute a image within $1000\times1000\times3$ pixels using $60,000$ painting actions, with an observation of size $41\times82\times3$.
We tested the run-time algorithm on machines with GPU and without GPU; both types of machines can compute the output within $300 secs$. 

\vspace{-0.1in}

\subsection{Application}
\begin{figure}[t]
\centering
\includegraphics[width=0.48\textwidth]{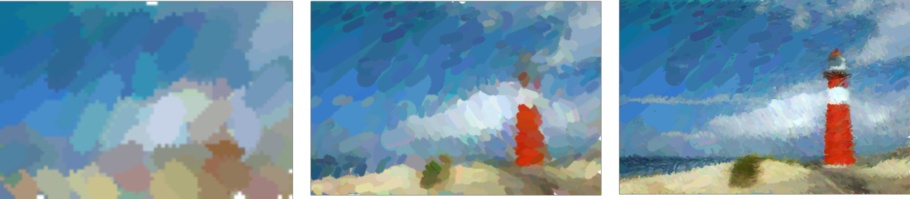}
\caption{\emph{Applications in different scales} The figure demonstrate how agents behave in different scales of the reference image. We up-sample the reference image and apply the trained painting agent to it. As a result, the framework gradually moves to a higher resolution reference image to let the agent draw details. Later scales paint directly on the canvas from previous scales.
}
\label{fig:scale}
\vspace*{-0.1in}
\end{figure}

After we have the trained the painting agent, we apply it at different image scales of the reference image. As shown in Fig.\ref{fig:scale}, we let the agent start with a low resolution reference image in order to quickly paint the high-level visual structure, and gradually move to a high resolution reference image to let the agent draw details. Note that later scales paint directly on the canvas from previous scales. 

Moreover, the learned policy is highly dependent on the training set of reference images. The painting agent can learn to paint the reference image in the implicit style of the training set. To generate the style transfer result, we collected paintings from different artists in different painting styles as different agents. Then we apply the learned painting agent to the unseen data. Fig.\ref{fig:pissaro} demonstrate how the pointillism style is  the implicit style from the training dataset to the reference image.


\begin{figure}[t]
\centering
\includegraphics[width=0.49\textwidth]{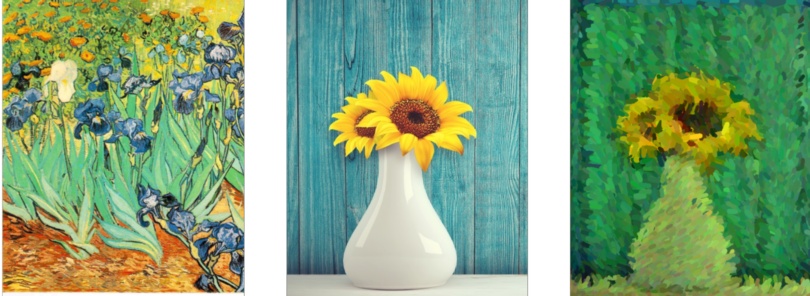}
\caption{\emph{Failure Cases:} 
We use (left image) as the training dataset  for our painting agent and apply it to reference image (middle)  and generate the final result (right). 
This example highlights the failure cases of our approach including: (1) off color, (2) not natural patterns, (3) changing color with a single stroke.
(1) is caused by the bias in the training dataset. 
(2) shows that the agent can paint across the edge of the shapes.
(3) shows that the agent may change color within a single stroke. These problems are caused by the  representation of the action. 
}
\label{fig:failure}
\vspace*{-0.1in}
\end{figure}

\begin{figure*}[ht]
\centering
\includegraphics[width=0.8\textwidth,]{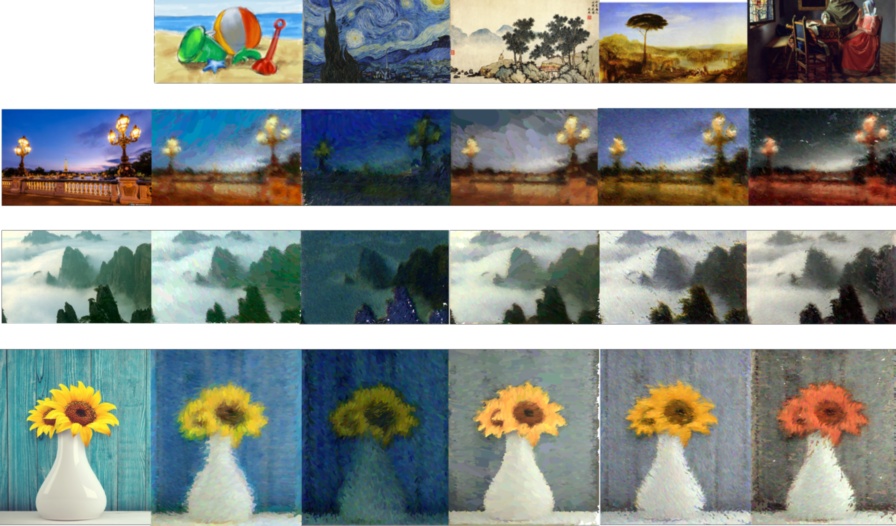}
\put(-380,-10){(a)}
\put(-315,-10){(b)}
\put(-250,-10){(c)}
\put(-180,-10){(d)}
\put(-100,-10){(e)}
\put(-45,-10){(f)}

\caption{{\em Style Transfer:} We trained the images using the data from the paintings in a specific styles.  The first row of each column (b-e) is the source image of the training dataset.  Our results show (a) reference images; (b) results generated by the agent trained by paintings in watercolor; (c) results generated by the agent trained by painting Starry Night by Van Gogh; (d) results generated by the agent trained by paintings by Shen Zhou; (e) results generated by the agent trained by paintings by Turner; (f) results generated by the agent trained by paintings by Vermeer.}
\label{fig:style}
\vspace*{-0.1in}
\end{figure*}

As shown in Fig.\ref{fig:failure},  our approach has some possible failure cases. These include off-color failure cases, which are caused by the bias in the training dataset. When our algorithm is trained with a limited number of reference images, the agent tends to choose a color that is closest to the training dataset. As a result, it is crucial to select the training dataset to make sure the colors are distributed in the entire color space. Sometimes we make use of the effect to generate a specific style of images as shown in Fig.\ref{fig:style}. 

Other cases are related to the action representation. In our setup, a stroke may include a bunch of segments. When the agent changes the configurations like color, angle and stroke width within the stroke, there is no pre-defined penalty or restriction to prevent it. Thus, some strokes may not look natural. However, the action representation with fewer restrictions can make the agent generate visual effects without complex simulation (e.g. dissolving effects in watercolor). 

\begin{figure*}[ht]
\centering
\includegraphics[width=0.76\textwidth,]{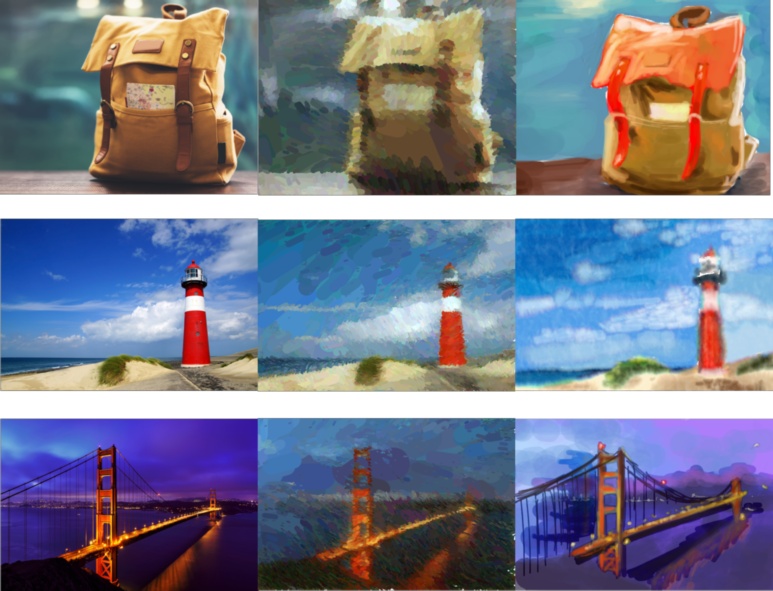}
\put(-350,-10){(a)}
\put(-200,-10){(b)}
\put(-70,-10){(c)}

\caption{ \emph{Our results compared with the human artwork:} We compare the result computed by our painting agent with the human artist painting. (a) are the reference images. (b) are generated by our painting agent, using watercolor image dataset in Fig.\ref{fig:data}. (c) are painted by human artists.}
\label{fig:update}
\end{figure*}

\begin{figure*}[ht]
\centering
\includegraphics[width=0.8\textwidth,]{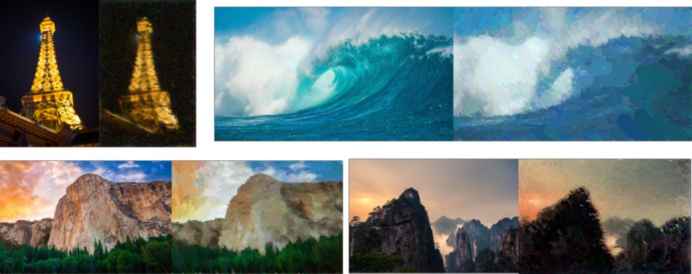}
\put(-370,67){(a)}
\put(-330,-10){(b)}
\put(-150,70){(c)}
\put(-110,-10){(d)}
\caption{{\em Result generated using PaintBot:} We trained the images  using specific styles. For each pair, the left is the reference image and the right is our result. We different training data corresponding to (a) Turner's paintings; (b) Watercolor paintings; (c) Watercolor paintings; (d) Vermeer's paintings.}
\label{fig:style1}
\vspace*{-0.12in}
\end{figure*}

%% file: conclusion.tex
\section{Conclusion, Limitations and Future Work}
We present an approach for training a reinforced natural media painting agent with limited training data. 
We incorporate a novel reinforcement learning framework into the problem and generate results in  a high dimensional a and continuous action space.
To reduce the search space of the policy, we design several techniques like curriculum learning and $\epsilon$-greedy sampling.
We highlight the performance on various reference images with different styles and resolutions.
We also demonstrate other applications of our approach like style transfer.  

Our approach has some limitations.
In our current implementation, we sample a small patch as the observation, which implies that the state representation contains limited information. 
Due to the computation complexity of the policy gradient algorithm, we use a small patch of size $41 \times 82 \times 3$. This can potentially make the policy network fall into a local minima. 
Another limitation is due to the choices of the action space. Although we use a 6-dimensional continuous action space, it is still can not represent all natural media painting styles. It may looks rigid sometimes especially when the agent makes a sharp turn while painting a long stroke.  Our current approach is mainly based on using  reinforcement learning for the painting problem, and many standard limitations of current reinforcement learning methods related to policy representation and choice of reward functions are also applicable. 

There are many avenues for future work. In addition to overcoming these limitations, we would like to evaluate our approach on more samples. We can use evaluate other rendering engines and different methods to train the painting policy network.
